\documentclass[journal]{IEEEtran}
\ifCLASSINFOpdf
\else
\fi
\label{begins}

\usepackage{graphicx}
\usepackage{subfig}
\usepackage{cite}
\usepackage{bm}
\usepackage{algorithm}
\usepackage[noend]{algpseudocode}
\usepackage{multicol,multirow,booktabs}
\usepackage{makecell}
\usepackage{amsmath}
\usepackage{longtable}
\usepackage{rotating}
\usepackage{caption}
\usepackage{amssymb}
\usepackage{pifont}
\usepackage{array}
\usepackage{color}

\begin{document}

%
\title{Domain Adaptive Adversarial Learning Based on Physics Model Feedback for Underwater Image Enhancement}
%
%
%

\author{Yuan~Zhou,~\IEEEmembership{Senior~Member,~IEEE,}
        ~and~Kangming~Yan
\thanks{Yuan Zhou is with the School of Electrical and Information Engineering,
Tianjin University, Tianjin 300072, China.}
\thanks{Kangming~Yan is with Tianjin International Engineering Institute, Tianjin University, Tianjin 300072, China.}}

%
%

\markboth{IEEE Transactions on Image Processing,~Vol.~XXX, No.~XXX}%
{Shell \MakeLowercase{\textit{\emph{et al.}}}: Bare Demo of IEEEtran.cls for IEEE Journals}
%



\maketitle

\begin{abstract}
\label{sec:0.abstract}
  Owing to refraction, absorption, and scattering of light by suspended particles in water, raw underwater images suffer from low contrast, blurred details, and color distortion. These characteristics can significantly interfere with the visibility of underwater images and the result of visual tasks, such as segmentation and tracking. To address this problem, we propose a new robust adversarial learning framework via physics model based feedback control and domain adaptation mechanism for enhancing underwater images to get realistic results. A new method for simulating underwater-like training dataset from RGB-D data by underwater image formation model is proposed. Upon the  synthetic dataset, a novel enhancement framework, which introduces  a domain adaptive mechanism as well as a physics model constraint feedback control, is trained to enhance the underwater scenes. Final enhanced results on synthetic and real underwater images demonstrate the superiority of the proposed method, which outperforms nondeep and deep learning methods in both qualitative and quantitative evaluations. Furthermore, we perform an ablation study to show the contributions of each component we proposed.
\end{abstract}


%
\IEEEpeerreviewmaketitle

\section{Introduction}
\label{sec:1.Introduction}
%
%
%
%

\IEEEPARstart{W}{ith} the development of science and technology, humans' sight has not been limited to the terrestrial area visible to the naked eye. The ocean, which occupies 71\% of the earth's surface area, is one of the hottest areas of exploration at this stage. Recently, underwater images have become the most effective tools to explore this treasure trove of resources and have emerged in a wide spectrum of aquatic applications, such as  deep ocean exploration, inspection of underwater infrastructures \cite{1} and cables \cite{2}, sea life monitoring \cite{3}, archeology \cite{4}, and control of underwater robotics \cite{5}.

Different from common images, underwater images suffer from poor visibility such as low contrast, color casts, and blurred details, resulting from the attenuation of the propagated light, mainly due to wavelength-dependent light absorption and scattering as well as the effects of low-end optical imaging devices \cite{6} \cite{7}.  The scattering and absorption attenuate the direct transmission and introduce surrounding scattered light. The attenuated direct transmission leads the intensity from the scene to be weaker and introduces color casts, while the surrounding scattered light causes the appearance of the scene to be washed out. In nature, this  distortion magnitude of  scattering and attenuation is extremely non-linear, and is affected by various factors, including the number of underwater particles, time of day, operational depth, overcast versus sunny, and imaging devices \cite{8}. Some examples of underwater images with different environmental conditions are illustrated in Fig. 1. Due to serious degeneration, it is hard to recover the realistic  color and appearance of underwater images, while color and appearance are crucial for underwater vision tasks \cite{9}, such as classification, detection, tracking, to name a few. Thereby, developing an effective solution to enhance contrast and restore color for these images is desirable.

\begin{figure}[t]
\label{fig:1}
\begin{center}
\includegraphics[scale=0.42]{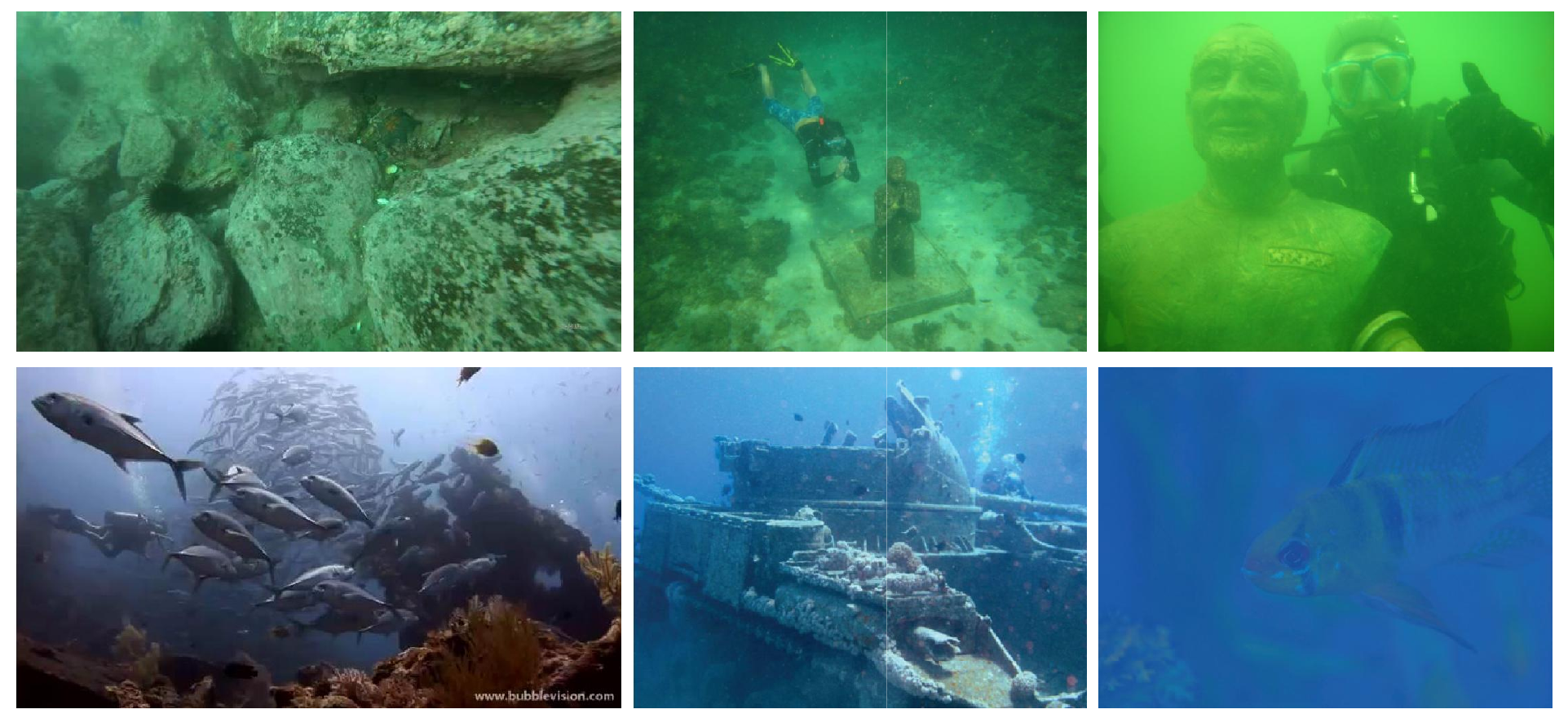}
\end{center}
\caption{Samples of underwater image with natural and man-made artifacts displaying the diversity of distortions that can occur. With the varying camera-to-object distances in the images, the distortion and loss of color varies between the different images.}
\end{figure}

Various processing methods for images degraded by the underwater environment have been developed in the past years. Traditional methods include image restoration and image enhancement. For image restoration methods \cite{10,11,12,13,14,15}, the degradation model of underwater optical imaging is  taken into consideration for reconstructing the images. Most of these methods are difficult to simulate and recover the complex underwater imaging process by estimating only a few parameters. They can only alleviate the color casts and blur effect of underwater images to a certain extent. While image enhancement techniques \cite{6,16,17,18} focus on adjusting image pixel values to acquire satisfactory results without depending on imaging model, they can only generate a single enhancement effect on various underwater images globally regardless of image style and imaging process. In addition, due to the lack of abundant training data, these traditional methods display poor generalization performance in different underwater images, and some of the results tend to be over-enhanced or under-enhanced.

Alternatively, Deep neural networks have been shown to be powerful non-linear function approximators, especially in the field of vision. For low-level vision tasks, e.g., image super-resolution \cite{19}, image de-raining \cite{20}, image de-noising \cite{21}, and de-hazing \cite{22}, powerful supervised learning models have obtained convincing success. Generally, these networks require large amounts of labeled data. However, images obtained in harsh and complex underwater scenario lack ground truth, which is a major hindrance towards adopting a similar supervised approach for correction. Some researchers try to solve the problem of the lack of ground truth in underwater images (by synthesizing paired underwater images from in-air data or proposing new weakly supervise constraint), and perform underwater image enhancement tasks through deep learning methods \cite{23,24,25,26,27,28,29}. However, underwater images synthesized by existing algorithms such as through physical models have a single style, which have visual differences and inter-domain differences with diverse real-world underwater images. Simultaneously, due to the complexity of the underwater image enhancement problem, e.g., the solution space of the corresponding problem is too large, a simple feed-forward network or generator with a random initialization is unable to estimate the solution well, and the weakly supervised constraint or only the adversarial loss does not ensure that the contents of the outputs are consistent with those of the inputs. So these algorithms are still less effective enough, it is still necessary to develop underwater image synthesis and enhancement methods for superior underwater visual quality and improve the performance of high-level vision tasks.

In this paper, we design an end-to-end solution for the complex and nonlinear underwater image formation procedure. More elaborately, a domain adaptation based and physical model constrained novel adversarial learning \cite{30} architecture is proposed, it was trained on the underwater scene prior based synthesized image pairs and real-world underwater images. Extensive experiments validate the superior performance of proposed framework with respect to robustness, flexibility and realistic for diverse water type. The main contributions of this paper are summarized as follows:

\begin{enumerate}
  \item
  \emph{We synthesized a novel image training set based on the physical imaging model in underwater scenario}, which is more in line with the visual effects of multiple real-world style underwater images.
  \item
  To the best of our knowledge \emph{we are the first that propose a physics model constrained learning algorithm so that it can guide the estimation in the GAN framework of underwater image processing}. The physics model acts as the \emph{feedback controller} of GAN based enhancement network, provides explicit constraints for this ill-posed problem, ensures that the estimated results should be consistent with the observed image and more realistic. The GAN with the physics model constrained learning algorithm is jointly trained in an end-to end fashion.
  \item
  Compared with the existing neural network based methods trained by synthesized underwater image pairs, to our best knowledge, \emph{this is the first attempt to introduce a domain adaptive mechanism to eliminate the domain gap between synthetic underwater images and real-world underwater images}, which helps the network trained on synthetic datasets also effective enough for enhancing real underwater images.
  \item
  Our method generalizes well both to synthetic and real-world underwater images with diverse color and visibility characteristics.
\end{enumerate}

The remainder of this paper is organized as follows: In Section \uppercase\expandafter{\romannumeral2}, we give a brief overview of the background knowledge and previous art. In Section \uppercase\expandafter{\romannumeral3}, The proposed method is described. Section \uppercase\expandafter{\romannumeral4} gives experimental results of our proposed methods and analyzes their effectiveness by comparing with previous works. We finally conclude this paper in Section \uppercase\expandafter{\romannumeral5}.

\section{Background Knowledge And Previous Art}
\label{sec:2.Related Work}

This section surveys the basic principles underlying underwater image formulation model and the function of domain adaptation, then reviews the main approaches that have been considered to restore or enhance the images captured under water.

\subsection{Underwater Image Formulation Model}
According to Jaffe-McGlamery imaging model \cite{31}, if the camera is not so far away from the scene, an underwater image can be regarded as a linear superposition of two components: 1) light which has not been scattered or absorbed in the intervening water, called the direct component; 2) light which enters the camera without reflection from the object, called backscatter. It can be formulated as follows:

\begin{equation}
\begin{split}
\label{1}
I(x)= J(x)\cdot t_{\lambda}(x) + &B_{\lambda}(1-t_{\lambda}(x)) ,\\
& \lambda\in \left\{red, green, blue \right\}
\end{split}
\end{equation}

where $I(x)$ is the captured underwater image; $J(x)$ is the clear latent image, also called as the scene radiance, that we aim to recover; $B_{\lambda}$ is the homogeneous global background light; $\lambda$ is the wavelength of the light for the red, green and blue channels; and x is a point in the underwater scene. The medium energy ratio $t_{\lambda}(x)$ represents the percentage of the scene radiance reaching at the camera after reflecting from the point $x$ in the underwater scene, which thereby causes color cast and contrast degradation. In other words, $t_{\lambda}(x)$ is a function of the wavelength of light λ and the distance $d(x)$ from the camera to the object surface:

\begin{equation}
\label{2}
t_{\lambda}(x)= Nrer(\lambda)^{d(x)},\lambda\in \left\{red, green, blue \right\}
\end{equation}

Where $Nrer(\lambda)$ is the normalized residual energy which is the ratio of residual energy to the initial energy per unit of distance and is dependent on the wavelength of light. For example, the bluish tone of the most underwater images is due to the fast attenuation of the red wavelength in open water as it possesses a longer wavelength than blue and green ones.

\subsection{Domain Adaptation}

Conventional machine learning algorithms rely on the assumption that the training and test data are drawn i.i.d. from the same underlying distribution. However, in practice it is common that there exists some discrepancy (domain gap) between training data and testing data. Domain adaptation aims to rectify this mismatch and tune the models toward better generalization at testing phase \cite{32,33,34,35}.

In Underwater image enhancement community and many other low-level tesks, various learning based methods call for synthetic dataset to train the enhancement models, generalizing to real-world underwater images. However, these methods ignored the domain gap between the synthetic training data and real-world testing data, which seriously affected the generalization ability of their model. To deal with this problem, we introduce a domain adaptive mechanism to eliminate the domain gap between synthetic underwater images and real-world underwater images, which helps the network trained on synthetic dataset also effective enough for enhancing real underwater images.

\subsection{Related work of Underwater Image Processing}
Given the importance of underwater vision, numerous methods towards improving underwater image quality have been proposed to address the degradation issues of underwater images. Generally, these algorithms can be categorized into three types including model-based restoration methods, model-free enhancement methods, and learning-based convolutional neural networks (CNNs).

Restoration method regards the recovery of underwater image as an inverse problem, which restores underwater images by estimating parameters of underwater image formation model. The dark-channel prior \cite{36} is the most commonly adopted prior, which is used in the estimation of the scene depth in a single image. some researchers apply this prior to process underwater images. For example, Drews et al. extended the classical DCP to underwater image restoration \cite{10}. Chiang et al. apply DCP and extend the work to compensate the attenuated light according to the scene depth and the normalized residual energy ratio in each light channel \cite{37}. Galdran et al. restored red channel to recover the lost contrast of un derwater images \cite{11}. Peng et al. adopted image blurriness with the image formation model to estimate the distance between the scene point and the camera, and thereby recovered underwater images \cite{14}. Moreover, Fattal \cite{38} proposed a novel method for single image dehazing, which takes advantage of a color-lines pixel regularity. Zhou et al. extended the color-lines model to underwater image restoration \cite{15} and demonstrated decent performance. However, many physical parameters and underwater optical property are required, making these restoration-based methods inflexible to be implemented in the harsh and complex real underwater environment.

Underwater image enhancement technologies always focus on adjusting image pixel values to produce a subjectively and visually appealing image. Iqbal et al. proposed an integrated color model and an unsupervised color correction method to enhance the visual quality of underwater images \cite{39} \cite{40}. Ancuti et al. proposed a fusion-based method to increase the contrast of underwater images and videos \cite{6}. Chani and Isa improved contrast and reduced noise of underwater images through modifying the integrated color model \cite{16}. Li and Guo proposed an underwater image enhancement method based on dehazing and color correction \cite{41}. Fu et al. proposed a simple yet effective retinex-based (RB) approach to enhance a single underwater image \cite{17}. Bianco et al. proposed a new color correction method for underwater imaging, which demonstrated the effectiveness of color correction in $L\alpha \beta$ color space \cite{42}.  However, These enhancement-based methods improve underwater scene contrast and image quality to some extent, but output images in some scenes become overenhanced or underenhanced, simultaneously their methods reckon without the underwater physical parameters.

Relying on abundant training data, deep learning \cite{43} techniques are capable of improving image quality in different underwater scenes. In virtue of the physical model, WaterGAN \cite{23} uses in-air images with corresponding depth information to generate the synthetic image for specific underwater scenarios. Li et al. develops a weak underwater image color correction model based on the cycle-consistent adversarial network (CycleGAN) \cite{44} and a multiterm loss function \cite{28}. Considering that CycleGAN can translate an image from one domain to another domain without paired training data or depth pairings, an underwater GAN (UGAN) \cite{24} employs it as a degradation process to generate paired training data, and then uses the model based on pix2pix \cite{45} to improve underwater image quality. However, due to the complexity of the underwater image quality improvment problem, e.g., the solution space of the corresponding problem is too large, a simple feedforward network with a random initialization is unable to estimate the solution well.

Different from previous methods, we propose an effective framework designed for underwater image enhancement with feedback control based on physical model. It trained by the synthetic underwater images, and a domain adaptive mechanism is introduced to eliminate the domain gap between synthetic training data and real-world underwater testing images, which would be demonstrated to be effective in ablation study. In addition, the proposed framework performs well in terms of both subjective and objective evaluations on synthetic underwater images and real underwater images.

\section{Methodology}
\label{sec:3 Methodology}

\begin{figure*}[hbt]
\label{fig:2}
\begin{center}
\includegraphics[scale=0.49]{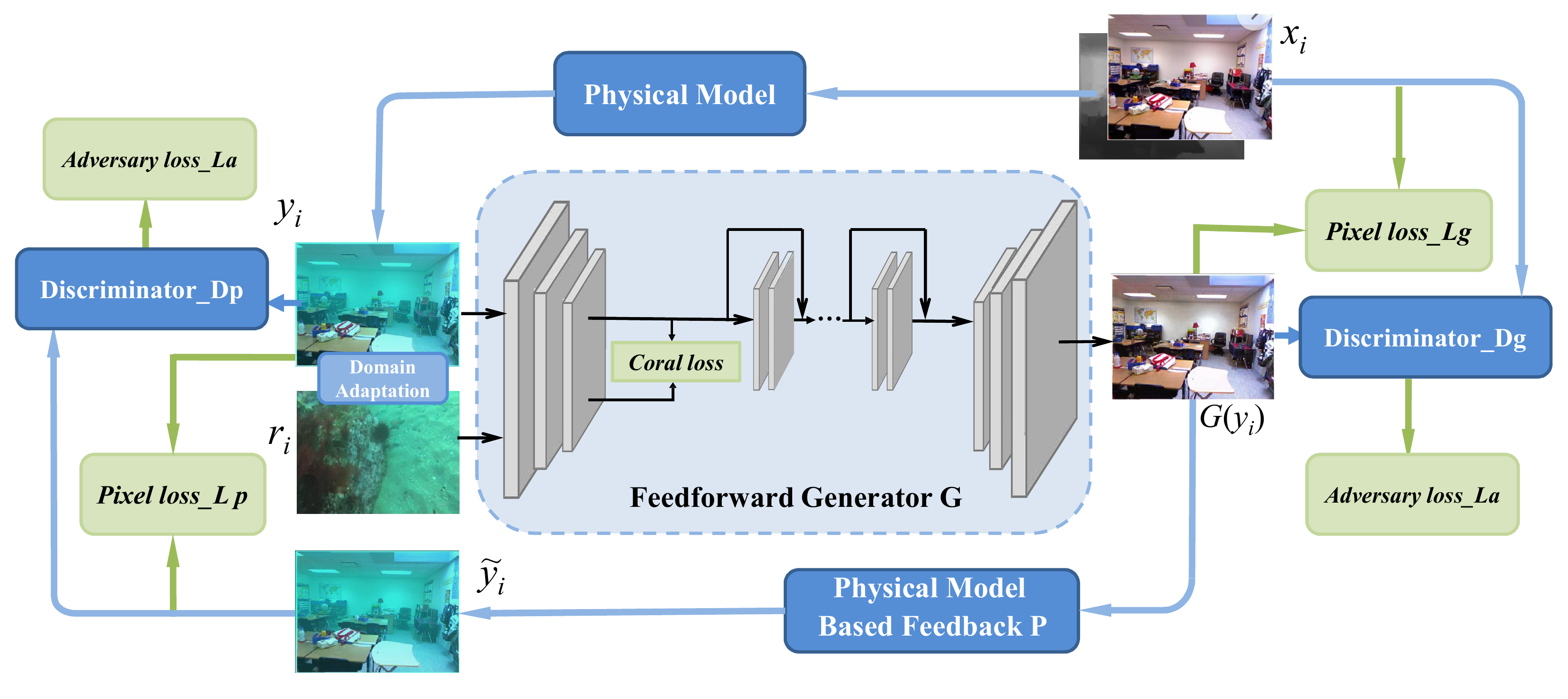}
\end{center}
\caption{Proposed framework. The discriminative network $D_{1}$ is used to classify whether the distributions of the outputs from the generator $G$ are close to those of the ground truth images or not. The discriminative network $D_{2}$ is used to classify whether the regenerated result $\widetilde{y_{i}}$ is consistent with the observed image $y_{i}$ or not. All the networks are jointly trained in an end-to-end manner.}
\end{figure*}

GAN \cite{30} have attracted favorable attention in machine learning community not only for its ability to learn the target probability distribution but also for its theoretically attractive aspects. Inspired by GAN and the priori knowledge of underwater image formulation model, we propose a novel framework to learn a end-to-end nonlinear mapping between the non-distorted image and the distorted image,  which can use the fundamental constraint to guide the training of GAN and ensure that the enhanced results are physically correct and seem real. We use the synthetic underwater images and its corresponding ground-truth to train the network. The synthesized underwater images can cover a variety of underwater scenario styles through reasonable randomization of parameters, but the real underwater scenes are extremely complex. So we introduce a domain adaptive mechanism in our network to eliminate the domain gap between synthetic training images and real-world underwater images, which helps the network trained on synthetic datasets also effective enough for enhancing real underwater images. As the flowchart shown in Fig. 2, the proposed model contains three main components, a feedforward generator network $G$, a feedback controller $P$, two discriminator networks $D_{g}$ and $D_{p}$. The networks have four types of term, including adversarial loss, cycle consistency loss, pixel loss, and domain adaptation loss (coral loss).

We first describe our method of synthetizing underwater training images in Section \uppercase\expandafter{\romannumeral3}-A, we then propose a novel underwater image enhancement framework in Section \uppercase\expandafter{\romannumeral3}-B. Finally, we show the optimization objective we used in Section \uppercase\expandafter{\romannumeral3}-C.

\subsection{Synthetizing Underwater Images}

To preserve the real color and content of the image, supervised methods are more suitable for underwater restoration. Unlike the high-level visual tasks \cite{46,47,48} where large training datasets with labels are often available, lacking underwater image dataset with corresponding ground truth constrains the development of deep learning-based underwater image enhancement and quality evaluation.  To solve this problem, we adopt an novel underwater image synthesis algorithm based on the underwater imaging physical model mentioned above and the observation of real underwater scenes. It was decided to simulate images based on the NYU dataset V2 \cite{49}. It’s relatively big, versatile and, most importantly, includes the ground truth depth information for each image, which is important for the method described later in this section. We modified this dataset to match the requirements of the abundant and various underwater scenes.

To convert images taken in the air into underwater styles, we apply Eqs. (1) and (2) to build three main types of underwater image datasets using the RGB-D NYU-v2 indoor dataset \cite{49} which consists of 1449 images ($J(x)$  of Eq.(1)) and corresponding depth information ($d(x)$ of Eq.(2)). For the parameter setting of normalized residual energy ($Nrer(\lambda)$) and homogeneous global background light information ($B_{\lambda}$) in the underwater image formation model, we made various attempts based on \cite{50} and \cite{51}, and compared the synthesis results with a large number of real underwater images of different scenes in terms of style and tone. Finally, we selected the three sets of parameter setting and randomization methods for $Nrer(\lambda)$ and $B_{\lambda}$  of red, green and blue channels for different water types, which is presented in Table 1 below. They cover different turbidity conditions from the clear waters in the offshore to the more turbid waters in the deep ocean, and the main tonal range of the underwater environment.

Through rational parameter setting and randomization, we synthesized a unified image training set contains different water types based on the physical imaging model in underwater scenario, which is more in line with the visual effects of multiple real-world underwater images, as shown in Fig.3. It is worth mentioning that, each of image pair in our synthetic dataset consists of four images, not only the synthetic underwater image and the its ground truth image taken in the air, but also the corresponding images containing background light information and the transmission map, which will be used in the feedback control module of our proposed enhancement framework. We selected 8900 pairs of images as our training set and all of them are resized to the canonical size of 256 $\times$ 256 pixels. To evaluate the effect of our proposed framework, we randomly select 80 images (keep the original size 480 $\times$ 640 ) from our synthetic image pairs, where the images have different shades and styles and are not used in the training stage. By the way, in addition to these synthetic data, we also compare with state-of-the-art methods using the commonly used real underwater images.

\begin{figure*}[t]
\label{fig:3}
\begin{center}
\includegraphics[scale=0.55]{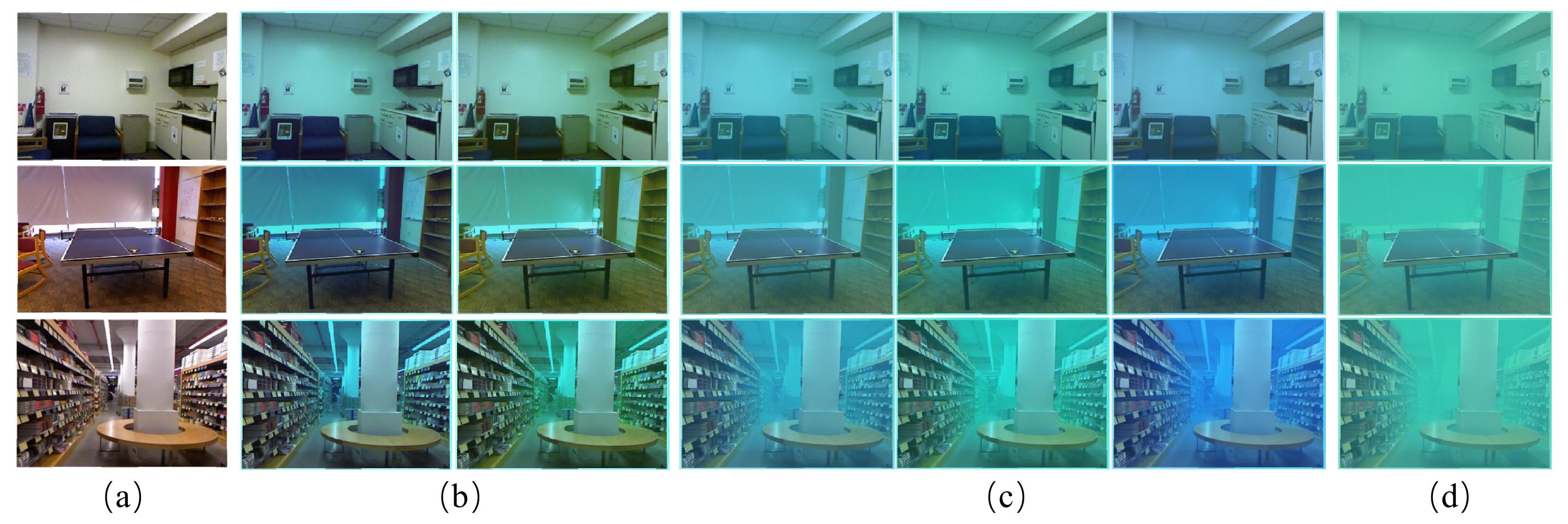}
\end{center}
\caption{The samples of synthesized underwater images from the NYU-v2 RGB-D dataset \cite{49} using a sample image and its depth map with the attenuation coefficients and background light shown in Table.1. (a) is the original image in NYU-v2 RGB-D dataset \cite{49}, while (b) (c) and (d) are the samples of synthetic underwater images with different degrees of degradation and  background lights.}
\end{figure*}

\begin{table*}[hbt]
\centering
\caption{
Three sets of parameter settings and randomization methods for $Nrer(\lambda)$ and $B_{\lambda}$  of red, green and blue channels for different water types.
}
\begin{tabular}{| c | c | c | c | c |}\hline
\textbf{Type}                & \textbf{Parameter}       & \textbf{Red}              & \textbf{Green}            & \textbf{Blue}             \\\hline
\multirow{2}{*}{(b)} & $Nrer(\lambda)$ & 0.79+0.06*rand() & 0.92+0.06*rand() & 0.94+0.05*rand() \\
                     & $B_{\lambda}$   & 0.05+0.15*rand() & 0.60+0.30*rand() & 0.70+0.29*rand() \\\hline
\multirow{2}{*}{(c)} & $Nrer(\lambda)$ & 0.71+0.04*rand() & 0.82+0.06*rand() & 0.80+0.07*rand() \\
                     & $B_{\lambda}$   & 0.05+0.15*rand() & 0.60+0.30*rand() & 0.70+0.29*rand() \\\hline
\multirow{2}{*}{(d)} & $Nrer(\lambda)$ & 0.67             & 0.73             & 0.67             \\
                     & $B_{\lambda}$   & 0.15             & 0.80             & 0.70\\\hline
\end{tabular}
\vspace{2mm}
\vspace{-2mm}
\label{table:1}
\end{table*}

\subsection{Proposed Enhancement Framework}

Our purpose is to obtain an effective and robust end-to-end underwater image enhancement framework for multiple water types. Through the network is trained through the synthetic underwater image training set mentioned above, it can be effectively used for the enhancement of real-world underwater images with different styles and scenarios, by which the color and content of subjects can be truly restored. The robustness of the network for multiple water types and the authenticity of the visual effects of the output results are the mainly remarkable characteristics of our method compared to other underwater image restoration and enhancement methods.

In order to make the network trained on synthetic dataset also effective enough for enhancing real underwater images, we introduce a domain adaptive mechanism to eliminate the domain gap between synthetic underwater images and real-world underwater images during the training procedure. Simultaneously, to give a physically correct result with more realistic visual effect for the ill-posed problem of underwater image enhancement, we add a feedback control module by the physics model constraint to the original end-to-end feedforward structure, which can guide the training of the generator. And to ensure the output of feedforward generator (i.e., $G(y_i)$) is consistent with the input $y_i$ under the model (1), we introduce an additional discriminative network. The proposed framework is shown in Fig.2. It contains one feedforward generative network $G$ with the domain adaptive mechanism, one feedback fundamental constraint $P$, and two symmetrical discriminative networks $D_{g}$ and $D_{p}$.

To the best of our knowledge, this is the first time that the knowledge of domain adaptation has been used in the underwater image processing algorithm to eliminate the domain gap between the training set and the test sets with a variety of real-world underwater scenes. It is also the first attempt to introduce a physical model based feedback control system into this research area. Both of them are significant contributions for the development of underwater optical image processing.

\subsubsection{Feed-forward Generative Network $G$}

Our generator is an end-to-end feedforward network whose purpose is to convert the input low-level underwater image into a processed clear image as output. The impressive performance of the image-to-image translation method such as \cite{44} encourages us to explore the similar generator structure. The generator G based on a forward CNN is an encoder-decoder structure \cite{52}, which is composed of residual blocks. It consists of three sections: the down-sampling feature extraction module, a feature-preserving reconstruction module and an up-sampling image reconstruction module. By means of a nine-residual-block stack, the downsample-upsample model learns the essence of the input scene, and a synthesized version will emerge at the original resolution after the de-convolution operations. The detailed network parameters are shown in Table.2.


\begin{table*}[!t]
\label{table:2}
\caption{
Network parameters. "CINR" denotes the convolutional layer with the instance normalization (IN) and ReLU;
"ResBlock" denotes the residual block which contains two convolutional layers with the IN and ReLU;
"CTINR" denotes the fractionally-strided convolutional layers with IN and ReLU;
"CINLR" denotes the convolutional layer with the IN and LeakyReLU.
}
\vspace{-2mm}
\centering
\begin{tabular}{lccc|cc|ccc}
\multicolumn{9}{c}{Parameters of the generative network}           \\
\toprule
Layers                 & CINR$_1$ & CINR$_2$ & CINR$_3$ & \multicolumn{2}{c|}{ResBlock$_1$-ResBlock$_9$} & CTINR$_1$ & CTINR$_2$ & CINR$_4$ \\
\hline
Filter size            &  7  &  3  & 3           & 3              & 3                       &  3   & 3    &  7   \\
Filter numbers         & 64  & 128 & 256         & ~~~~~256~~~~~  &~~~~~256~~~~~            & 128  & 64   &  3   \\
Stride                 & 1   & 2    & 2          & 1              & 1                       & 2  & 2  &  1    \\
\bottomrule
\vspace{-1mm}
\end{tabular}
\begin{tabular}{lccccc}
\multicolumn{6}{c}{Parameters of the discriminative network}           \\
\toprule
Layers                 & CINLR$_1$ & CINLR$_2$ & CINLR$_3$ & CINLR$_4$ & CINLR$_5$ \\
\hline
Filter size            &  4  &  4  & 4           & 4               &  4   \\
Filter numbers         & 64  & 128 & 256         & 512             & 1 \\
Stride                 & 2   & 2    & 2          & 1               & 1 \\
\bottomrule
\end{tabular}
\end{table*}

As mentioned above, in order to make the network trained on synthetic datasets also effective enough for enhancing real underwater scenario, we introduce a domain adaptive mechanism to eliminate the domain gap between synthetic underwater images and real-world underwater images. During the training procedure, our generator $G$ take the synthetic underwater images $y_i$ and randomly selected real-world underwater images $r_i$ as input, while the output is only the clear images $G(y_i)$ without “water tune” corresponding to the input synthetic underwater image $y_i$. In the down-sampling module, features of unpaired synthetic underwater image $y_i$ and real-world underwater image $r_i$ are extracted, represented by $F(y_i)$ and $F(r_i)$, respectively. In general, both $F(y_i)$ and $F(r_i)$ are 3D tensors, whose size is $c\times h\times w$. For the sake of analysis, we consider the feature tensor as set of $(h\times w)$ $c$-dimensional local descriptors. Furthermore, local descriptors of synthetic underwater image and real-world underwater image are regard as the source and target domain, respectively. By the constrain adding to the end of down-sampling module which will explained in detail in Section \uppercase\expandafter{\romannumeral3}-C, the feature extraction process can be guided to aligns the second-order statistics of the source and target distributions. It will help the features of the synthetic training data obtained by down-sampling module similar to the feature representations of real underwater images, eliminating their inter-domain differences. Then, followed by the feature-preserving reconstruction module consist of nine residual blocks and the up-sampling image reconstruction module to generate output images from the features of the synthetic underwater images.

\subsubsection{Feed-back Control $P$}
We note that the CNN or GAN based methods with the observed data $y$ as the input has shown promising results in underwater image enhancement\cite{23,24,25,26,27,28,29}. However, these methods does not guarantee whether the solutions satisfy the physics model (1) or not and thus fails to generate clear and real enough images as illustrated in Section I. In this work, we develop a new method to improve the estimation result of GAN under the guidance of the physics model (1).

The feedforward generative network mentioned above learns the mapping function $G$ and generates the intermediate enhanced image $G(y_i)$ from the input $y_i$. Then, we apply the physics model (1) to $G(y_i)$ to get the regenerated underwater style image $y_i$ :

\begin{equation}
\begin{split}
\label{3}
\widetilde{y_i}=G(y_i)\cdot t_{i_{\lambda}}+B_{i_{\lambda}}(1-t_{i_{\lambda}}) , \lambda\in \left\{red, green, blue \right\}
\end{split}
\end{equation}
Where $t_{i_{\lambda}}$ donate the transmission map and and $B_{i_{\lambda}}$ is the background light which is only used in the training process. Note that the transmission map $t_{i_{\lambda}}$ and the background light $B_{i_{\lambda}}$  in Eq.(3) is known, which is also used to generate underwater style image $y_i$ from original clear image $x_i$ when synthesizing the training data. As mentioned in Section \uppercase\expandafter{\romannumeral3}-A, each of image pair in our dataset consists of four images, not only the synthetic underwater image $y_i$   and the its ground truth image $x_i$  taken in the air, but also the corresponding images $B_{i_{\lambda}}$ and $t_{i_{\lambda}}$ which containing background light and the transmission map information.

This physics model based module acts as the feedback controller of GAN based enhancement network, provides explicit constraints for this ill-posed problem, ensures that the estimated results should be consistent with the observed image and seem more realistic. We note that although the proposed framework is trained in an end-to-end manner, it is constrained by a physics model and thus is not fully blind in the training stage. With the learned generator $G$, the test stage is blind, we can directly obtain the final results by applying it to the input real underwater images.

\subsubsection{Discriminators}

 Our discriminator is modeled as a PatchGAN \cite{44,52} , which discriminates at the level of image patches with fewer parameters than a full image discriminator and achieve state-of-the art results in many vision problems. As opposed to a regular discriminator, which outputs a scalar value corresponding to real or fake, our PatchGAN discriminator outputs a 32$\times$32$\times$1 feature matrix, which provides a metric for high-level frequencies. The parameters of the discriminative network is shown in Table 2.

 There are two discriminators with the same structure in our proposed framework. The discriminative network $D_g$ takes the ground truth $x_i$ and the intermediate enhanced images $G(y_i)$ as the inputs and it is used to classify whether $G(y_i)$ is clear or not. The other discriminative network $D_p$ takes the synthetic underwater image $y_i$ and the regenerated image $\widetilde{y_i}$ as the inputs, which is used to classify whether the generated results satisfy the physical model or not.

\subsection{Optimization objective}

The fundamental GAN algorithm learns a generative model via an adversarial process. It simultaneously trains a generator network and a discriminator network by optimizing:

\begin{equation}
\begin{split}
\label{4}
\min \limits_G \max \limits_D &E_{x\sim P_{data} (x)} \left[ \log D(x) \right]\\
+ &E_{z\sim P_{z} (z)} \left[ \log (1- D(G(z))) \right]
\end{split}
\end{equation}
where $z$ donates random noise, $x$ donates a real image, and $D$ denotes a discriminator network, $G$ denotes the generator network. In the training process of GAN, the generator generates samples (i.e., $G(z)$) that can fool the discriminator, while the discriminator learns to distinguish the real data and the samples from the generator.

However, the contents of the generated images only based on this training loss may be different from the ground truth images. The supervise information provided by the discriminator and adversary loss along is not strong enough. To ensure that the contents of the generated results from the generative networks are sufficient close to those of the ground truth images and also consistent with those of the inputs under the underwater image physical formulation model (1), we use the $L_1$ norm regularized pixel-wise loss functions:
\begin{equation}
\begin{split}
\label{5}
L_g = \sum_i\|G(y_i) - x_i\|_1
\end{split}
\end{equation}
and

\begin{equation}
\begin{split}
\label{6}
L_m = \sum_i\|\widetilde{y}_i - y_i\|_1
\end{split}
\end{equation}
so

\begin{equation}
\begin{split}
\label{7}
L_{pixel} = \frac{1}{2} \cdot (L_g + L_m)
\end{split}
\end{equation}
in the training stage. To make the generative network learning process more stable, we further use the loss function:
\begin{equation}
\begin{split}
\label{8}
L_{cycle} = \sum_i\|G(\widetilde{y}_i) - x_i\|_1
\end{split}
\end{equation}
to regularize the generator G.

Finally, we propose a new objective function

\begin{equation}
\begin{split}
\label{9}
L_a = \sum_i [\log D_g(x_i)] + [\log (1-D_g(G(y_i)))] \\
+ [\log D_p(y_i)] + [\log (1-D_p(\widetilde{y}_i)]
\end{split}
\end{equation}
to ensure that the output of GAN is consistent with the observed input under the underwater image formulation model (1).

Basing the former objective functions, our network is able to learn powerful representations from large quantities of synthetic underwater images with ground truth, however, it can not guarantee the model generalize well across changes in input distributions while testing the result of the network by the real underwater images, for the reason that the training and testing data are not independent and identically distributed (i.i.d.). So as mentioned in Section \uppercase\expandafter{\romannumeral3}-B, we introduce a domain adaptive mechanism into our framework to compensate for the degradation in performance due to domain shift.

In this work, we use a differentiable loss function named CORAL loss \cite{53} by incorporate it directly into the down-sampling feature extraction module of our feedforward generative network, which can minimizes the difference in second-order statistics (covariances) between synthetic underwater image $y_i$ (source) and real-world underwater image $r_i$ (target) features:
\begin{equation}
\begin{split}
\label{10}
L_{coral} = \frac{1}{4d^2} \|C_S - C_T\|_F ^2
\end{split}
\end{equation}
where $\|\cdot \|_F ^2$ denotes the squared matrix Frobenius norm, and $C_S$ ($C_T$ ) denote the feature covariance matrices of source domain and target domain, $d$ donates the number of channels for the features. The covariance matrices of the source and target data are given by:
\begin{equation}
\begin{split}
\label{11}
C_S = \frac{1}{n_S -1} (F_y ^\mathsf{T} F_y - \frac{1}{n_S}(\textbf{1}^\mathsf{T} F_y)^\mathsf{T}(\textbf{1}^\mathsf{T} F_y))
\end{split}
\end{equation}
\begin{equation}
\begin{split}
\label{12}
C_T = \frac{1}{n_T -1} (F_r ^\mathsf{T} F_r - \frac{1}{n_T}(\textbf{1}^\mathsf{T} F_r)^\mathsf{T}(\textbf{1}^\mathsf{T} F_r))
\end{split}
\end{equation}
where $\textbf{1}$ is a column vector with all elements equal to 1. $n_S$ and $n_T$ represent the number of local descriptors $(h\times w)$ of source image (synthetic underwater image $y_i$) and target image (real-world underwater image $r_i$), respectively. $F_y$ and $F_r$ are the feature matrices of synthetic underwater image and real-world underwater image generated by the down-sampling feature extraction module.

By optimizing the coral loss embedded in the generator, the feature of synthetic underwater image training set and real-world underwater images can be mapped to a common feature subspace, eliminate their inter-domain differences.

Finally, the total optimization objective we proposed is the linear combination of the abovementioned losses with weights as follows:
\begin{equation}
\begin{split}
\label{13}
L_{loss} = L_{a} + \lambda _1 L_{cycle}+\lambda _2 L_{pixel}+\lambda _3 L_{coral}
\end{split}
\end{equation}
where $\lambda _1$, $\lambda _2$ and $\lambda _3$ are weight parameters.

\begin{figure*}[htpb]
\label{fig:4}
\begin{center}
\includegraphics[scale=0.40]{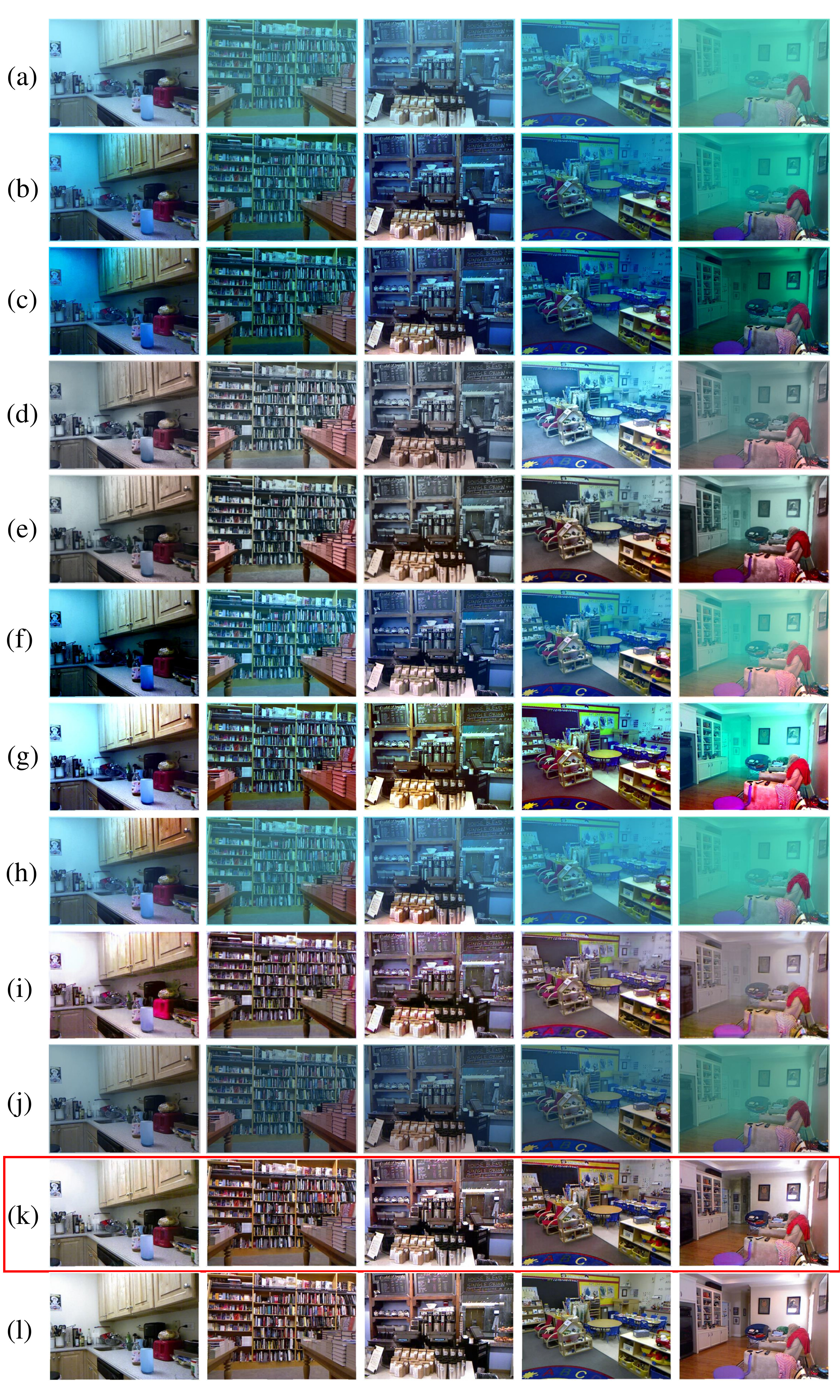}
\end{center}
\caption{Qualitatively comparison between our methods with contemporary approaches in terms of processing quality for samples from synthetic underwater image test set. (a) Raw synthetic underwater images. (b) Results of DCP \cite{36}. (c) Results of UDCP \cite{10}. (d) Results of Fusion \cite{6}. (e) Results of RB \cite{17}. (f) Results of IBLA \cite{14}. (g) Results of UCL \cite{15}. (h) Results of DehazeNet \cite{22}. (i) Results of Cyclegan \cite{44}. (j) Results of DUIENet \cite{25}. (k) Results of our method. (l) Ground truth. The types of underwater images in the first column are with different degeneration degrees and background light. The results illustrate that our method removes the light absorption effects and recovers the original colors without any artifacts.}
\end{figure*}

\section{Experiments}
\label{sec:4 Experiments}

In this part, we perform qualitative and quantitative comparisons with the state-of-the-art underwater image enhancement methods on both synthetic and real-world underwater images.  These compared methods include the DCP method \cite{36}, the underwater DCP method \cite{10}, the improved retinex-based (RB) method \cite{17}, the image blurriness and light absorption (IBLA) based method \cite{14}, the fusion enhance (Fusion)  method \cite{6}, the color-line based underwater image restoration (UCL) method \cite{15}, Dehazenet \cite{22}, Cyclegan \cite{44} and DUIENet \cite{25}, thanks to their representativeness in single image de-hazing, tradition underwater image restoration and enhancement, and deep learning based image style transfer or underwater image enhancement, respectively. We run the source codes provided by the authors with the recommended parameter settings to produce the best results for an objective evaluation.

To verify the performance of different methods, subjective and objective evaluations including quality metrics and user study are carried out. At last, we conduct an ablation study to demonstrate the effect of each component in our framework. To validation the generalization ability of our proposed method for different underwater scenarios, we collected a test set of 80 real-world underwater images acquired from \cite{6,37} and the internet, these images have obvious characteristics of underwater image quality degradation (e.g., color casts, decreased contrast, and blurring details) and are taken in a diversity of underwater scenes. Some of the testing images are shown in Fig.1. And the original underwater images presented in this paper are extracted from this collected dataset.

\subsection{Training Details}
\label{sec:4.1 Training Details}

In our training process, The training data set consists of 8900 pairs of images with a resolution of 256$\times$256. We train the models using the Adam optimizer \cite{54} with an initial learning rate 0.0002, $\beta _1$ to 0.50, $\beta _2$ to 0.999. We set the batch size to be 1. After obtaining generator $G$, as we know the paired training data$\left\{x_i,y_i \right\}$ and the corresponding physics model parameters(the transmission map $t_{i_{\lambda}}$ and the background light $B_{i_{\lambda}}$) that are used to synthesize $y_i$ from $x_i$, we apply the same physics model parameters to $G(y_i)$ and generate $\widetilde{y}_i$. Then the discriminator $D_g$ takes the ground truth $x_i$ and $G(y_i)$ as the input while the other discriminative network $D_p$ takes $y_i$ and $\widetilde{y}_i$ as the input. We implemented our network with the PyTorch framework and trained it using 64 GB memory and 11 GeForce Nvidia GTX 1080 Ti GPU for 60 epochs.

\begin{figure*}[htpb]
\label{fig:5}
\begin{center}
\includegraphics[scale=0.66]{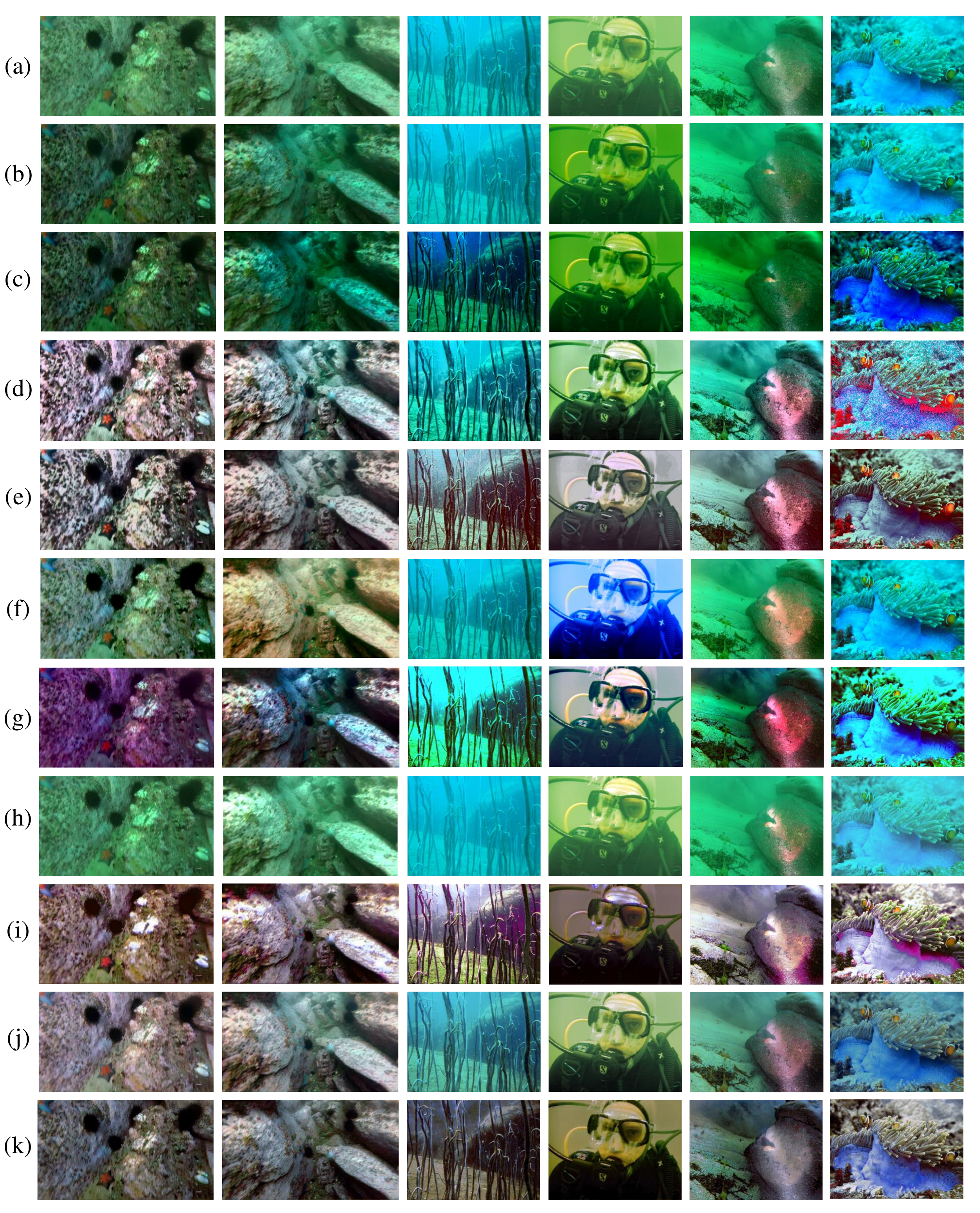}
\end{center}
\caption{Qualitatively comparison between our methods with contemporary approaches in terms of processing quality on real-world underwater images. (a) Raw synthetic underwater images. (b) Results of DCP \cite{36}. (c) Results of UDCP \cite{10}. (d) Results of Fusion \cite{6}. (e) Results of RB \cite{17}. (f) Results of IBLA \cite{14}. (g) Results of UCL \cite{15}. (h) Results of DehazeNet \cite{22}. (i) Results of Cyclegan \cite{44}. (j) Results of DUIENet \cite{25}. (k) Results of our method. The underwater images in the first row are with different degeneration degrees and background light. The results illustrate that our method produces the results without any visual artifacts, color deviations, and over-saturations. It also unveils spatial motifs and details(Best viewed on high-resolution display with zoom-in.)}
\end{figure*}

\subsection{Evaluation on Synthetic Underwater Image}
\label{sec:4.2 Evaluation on Synthetic Underwater Image}

\begin{table}[t]
\centering  
\label{table:3}
\caption{ Quantitative evaluations on test set. As seen, our method achieves the best scores in all metrics.}
\begin{tabular}{lccc}  
\hline
\textbf{Method} &\textbf{MSE} &\textbf{PSNR} &\textbf{SSIM}\\ \hline
Original &3864.3577 &12.6176 &0.7524\\
DCP &4604.2207 &11.7758	&0.7011\\
UDCP &5788.5313	&10.9260 &0.6553\\
Fusion &2528.0611 &14.5415 &0.8211\\
RB &1217.8306 &17.6973 &0.8521\\
IBLA &2291.2073	&15.3116 &0.7941\\
UCL &2035.3419 &15.9243	&0.8089\\
DehazeNet &4192.5230 &12.2191 &0.7355\\
Cyclegan &506.4429 &21.7324	&0.8707\\
DUIENet &2700.4965 &14.1571 &0.8254\\
Ours &\textbf{90.2791} &\textbf{28.8487} &\textbf{0.9532}\\
\hline
\end{tabular}
\end{table}
We first evaluate the result of underwater image enhancement of the proposed method using the synthesized underwater images as mentioned in Section \uppercase\expandafter{\romannumeral3}-A with nondeep and deep learning compared methods. The test dataset contains 80 synthetic images which includes multiple degradation degrees corresponding with their ground truth. Some of the subjective results of different methods are shown in Fig.4. As we can see, the DCP \cite{36}, UDCP \cite{10} and DehazeNet \cite{22} nearly fail to correct the underwater color meanwhile Fusion \cite{6}, RB \cite{17}, IBLA \cite{14}, UCL \cite{15}, and DUIENet \cite{25} introduce artificial color or color deviations obviously, although these methods eliminate the underwater color and degradation effect to a certain extent, they still retain the obvious underwater color style and haze blur effect. On the other hand, our method not only enhances the visibility of the images but also restores an aesthetically pleasing texture and vibrant yet genuine colors. In comparison to other methods, the visual quality of our results nearly the same as the ground-truth.

Furthermore, we quantify the accuracy of the recovered images on the synthetic test set including 80 samples for different degradation degree. In Table 3, the accuracy is measured by three different metrics: mean square error (MSE), peak signal to noise ratio (PSNR), and the structural similarity index metric (SSIM) \cite{55}. The quantitative results are obtained by comparing the results of each method with the corresponding ground truth image. In the case of MSE and PSNR metrics, the lower MSE (higher PSNR) denotes the result is closer to the ground truth in terms of image content. In the case of the SSIM metric, the higher SSIM scores mean the result is more similar to the ground truth in terms of image structure and texture. Here, the presented results are the average scores. The values in bold represent the best results.

Table 3 shows that the proposed method achieves the best performance in terms of all full-reference image quality assessment metrics against state-of-the-art methods, demonstrating its effectiveness and robustness. As we can see, regarding all the three metrics, our method is significantly better than the compared methods with absolute superiority.

\subsection{Evaluation on Real-world Underwater Image}
\label{sec:4.3 Evaluation on Real-world Underwater Image}

\begin{table}[t]
\centering  
\label{table:4}
\caption{  Underwater image quality evaluation of different processing methods on real-world underwater images. The best result is in bold.}
\begin{tabular}{lcccc}  
\hline
\textbf{Method} &UISM &UICM &UIConM &\textbf{UIQM}\\ \hline
Original &3.7823 &0.8239 &0.5607 &3.1449\\
DCP &4.0432	&1.5436	&0.7230	&3.8224\\
UDCP &3.7733 &2.0890 &0.7712 &3.9303\\
Fusion &5.1587 &3.7882 &0.8005 &4.4923\\
RB &4.8236 &3.2928 &0.7728 &4.2801\\
IBLA &3.9638 &2.9803 &0.5809 &3.3314\\
UCL &3.8997	&4.3608	&0.7272	&3.8748\\
DehazeNet &4.0402 &0.7002 &0.6164 &3.4167\\
Cyclegan &6.1765 &2.1636 &0.7036 &4.4006\\
DUIENet &4.5795	&3.0063	&0.6844	&3.8842\\
Ours &6.2318 &1.6609 &0.8264 &\textbf{4.8418}\\
\hline
\end{tabular}
\end{table}

\begin{table*}[t]
\centering  
\label{table:5}
\caption{  User study on real-world underwater image dataset. The best result is in bold.}
\begin{tabular}{cccccccccccc}  
\hline
Method &Original &DCP &UDCP &Fusion &RB &IBLA &UCL &DehazeNet & Cyclegan &DUIENet & Ours\\ \hline
Scores &4.0 &3.6 &3.9 &5.7 &6.2 &5.0 &4.8 &4.2 &5.6 &5.9 &\textbf{8.0}\\
\hline
\end{tabular}
\end{table*}

In this part, we evaluate the proposed method on real-world underwater images. The subjective evaluation with competitive methods, namely visual comparisons, is presented in Fig. 5. As we can see, the original real-world underwater images suffer from poor visibility, and all methods have some effects in improving the image visibility. But although the haze in the raw underwater images are removed by DCP, UDCP, IBLA and UCL, the visibility, color, and details are not good enough, their results still retain a distinct underwater color style with the greenish and bluish tone. The RB and Fusion methods, as typical image enhancement method does not take the underwater imaging formation model into consideration, the results have shown the lack of edges and details information when zoomed in, and these methods make images color distortion and noise, the color of the image does not conform the laws of nature.

For deep neural networks methods, the results of DehazeNet do not have a good vision, it cannot change the color and content of the turbid underwater images. This network was developed and trained for removing haze from the images taken in air, which is probably the reason for the very poor performance on the underwater images. The CycleGAN used here was retrained and fine-tuned by the synthetic underwater image training set, which is same with our method, so it works well in changing the overall scenario styles of underwater images. However, it makes the color of generated images changed, and the content and structure of turbid underwater images are slightly distorted.  As a recently proposed baseline model for underwater image enhancement, DUIENet removes the haze on the underwater images and remits color casts  quite effectively, but some of the results still retain a distinct underwater style, especially for inputs with a larger depth of field, which may affect the results of high level tasks. In contrast, our method shows promising results on all of the real-world images. The greenish and bluish tone is totally removed as if our results were taken on the ground, without introducing any artificial colors, color casts, over- or under-enhanced areas, which matches the nature underwater scenes. At the same time, it is obvious that the proposed algorithm can remove haze effect well, enhance the detailed information as clean as possible.

In order to make our method more convincing, we choose underwater image quality measure (UIQM) \cite{56} which is the non-referenced metric to evaluate the underwater image enhancement methods and restoration ones. This metric has three underwater image attribute measures: the underwater image colorfulness measure (UICM), the underwater image sharpness measure (UISM), and the underwater image contrast measure (UIConM). Each attribute is used to assess one aspect of the underwater image degradation. Therefore, the UIQM is given as follows:
\begin{equation}
\begin{split}
\label{14}
UIQM = c_1 \times UICM + c_2 \times UISM + c_3 \times UIConM
\end{split}
\end{equation}
where the colorfulness, sharpness, and contrast measures are linearly combined together. And the three parameters are $c_1$; $c_2$, and $c_3$. Their values are set to 0.0282, 0.2953, and 3.5753 according to the paper \cite{56}. Therefore, we believe that this metric can provide a comprehensive assessment of the effectiveness of various methods. Table 4 lists the average values obtained by different methods on out test set which contains 80 real-world underwater images. The best results of the final index (UIQM) is marked in bold. It can be seen that the UIQM of the proposed method is remarkably larger than the other methods.

For a more objective assessment, we conduct a user study to provide realistic feedback and quantify subjective visual quality. We randomly selected 30 real-world underwater images from our collected test set, which covers a diversity of underwater scenes, different characteristics of quality degradation, and a broad range of image content. We show samples from this dataset in Fig.1. And some corresponding results have been presented in Fig. 5. The results of different methods were randomly displayed on the screen and compared with the corresponding raw underwater images. After that, we invited 20 participants who had experience with image processing to score results. There was no time limitation for each participant. Moreover, the participants did not know which results were produced by our method. The scores ranged from 1 (worst) to 10 (best). As baseline, we set the scores of raw underwater images to 4.0, while one expects that the results with high contrast, good visibility, natural color, and authentic texture should receive higher ranks while the results with over-enhancement/exposure, under-enhancement/exposure, color casts, and artifacts should have lower ranks. The average subjective scores are given in Table 5. As we can see, our method receives the highest rankings, which indicates that our method can generates visually pleasing results.

\subsection{Ablation Study}
\label{sec:4.4 Ablation Study}

To demonstrate the effect of each component in our framework, we carry out an ablation study involving the following experiments:

(i)	Our method removes domain adaptation operation(-DA);

(ii)	Our method removes physical model based feedback control system(-PF);

(iii)	Our method removes pixel-wise losses(-PL).

\begin{table}[h]
\centering  
\label{table:6}
\caption{ Underwater image quality evaluation of different variants of the proposed method. The best result is in bold.}
\begin{tabular}{ccccc}  
\hline
  &UISM &UICM &UIConM &\textbf{UIQM}\\ \hline
-PL &6.5320	&1.5542	&0.7235	&4.5593\\
-PF &6.6536	&1.1653	&0.8093	&4.8911\\
-DA &6.6841	&1.6401 &0.8397 &5.0222\\
Ours &6.7606 &1.4928 &0.8528 &\textbf{5.0874}\\
\hline
\end{tabular}
\end{table}
We carry out the test on 50 real-world underwater images by quantitative evaluation. The average scores in terms of underwater image quality measure (UIQM) metric is reported in Table 6, and the best result is marked in bold. We notice that both physical model based feedback controller and domain adaptation mechanism could improve the final results of UIQM for the enhanced images.

\section{Conclusion}
\label{sec:5 conclusion}
In this paper, we presented an underwater image enhancement network inspired by underwater scene prior. Firstly, a new method for simulating underwater-like images which is more suitable for underwater scene has been proposed. Base on this, an adversarial learning architecture with domain adaptative mechanism and physical model constraint feedback control introduced in was trained to enhance the underwater images. Finally, numerous experiments are performed to demonstrate the superiority of the proposed method both on synthetic and real underwater images. In addition, the experimental results of ablation study also demonstrate that the physical model based feedback control and domain adaptation mechanism we proposed boost the performance quantitatively and qualitatively. Furthermore, our method can be used as a guide for subsequent research of the learning-based underwater image processing and similar low-level tasks such as image dehazing and super-resolution reconstruction.

\ifCLASSOPTIONcaptionsoff
  \newpage
\fi



%

%
%
\label{sec:reference}
\bibliographystyle{IEEEtran}
\bibliography{reference}

%








\end{document}